# Enhancing Automated Decision Support across Medical and Oral Health Domains with Semantic Web Technologies

Tejal Shah[1,2]
School of Computer Science & Engineering
University of New South Wales
Sydney, Australia
Email: tejals@cse.unsw.edu.au

Fethi Rabhi
School of Computer Science & Engineering
University of New South Wales
Sydney, Australia
Email: f.rabhi@unsw.edu.au

Pradeep Ray
[1]Asia-Pacific Ubiquitous Healthcare Research Centre
University of New South Wales
Sydney, Australia
Email: p.ray@unsw.edu.au

Kerry Taylor
[2]Computational Informatics
Commonwealth Scientific and Industrial Research Organisation
Canberra, Australia
Email: Kerry.Taylor@csiro.au

**Abstract**

*Research has shown that the general health and oral health of an individual are closely related. Accordingly, current practice of isolating the information base of medical and oral health domains can be dangerous and detrimental to the health of the individual. However, technical issues such as heterogeneous data collection and storage formats, limited sharing of patient information and lack of decision support over the shared information are the principal reasons for the current state of affairs. To address these issues, the following research investigates the development and application of a cross-domain ontology and rules to build an evidence-based and reusable knowledge base consisting of the inter-dependent conditions from the two domains. Through example implementation of the knowledge base in Protégé, we demonstrate the effectiveness of our approach in reasoning over and providing decision support for cross-domain patient information.*

## Keywords

OWL, SWRL, decision support, rules, ontologies

## INTRODUCTION

The ubiquitous presence of technology has seen healthcare delivery systems becoming increasingly computer based and more collaborative and integrated. For example, there are health information systems that record patient information during a medical visit and then interoperate with pathology, radiology and pharmaceutical systems to collect the respective information of the same patient thereby providing a more comprehensive picture for the healthcare practitioners to base their decisions on. However, such interoperability is negligible or very limited between medical and dental/oral health systems leading to silos within the medical and dental (specifically oral health) domains. The formation of silos further leads to fragmented delivery of care to patients who may need coordinated medical and oral health management. There is ample evidence in literature that several medical and oral health conditions are associated such that the progress of one affects the progress of the other. Hence it is imperative that inter-dependent medical and oral conditions should be managed simultaneously. For example, as shown in (Cutler et al. 1999), decreased metabolic control in diabetes mellitus type 2 has a negative impact on the periodontal health[1] of the patients and aggravates pre-existing periodontitis[2]. Periodontitis further adversely affects the prognosis of other medical conditions such as respiratory and cardiovascular diseases, stress, can affect the immune system leading to conditions associated with immune suppression, cause nutritional compromise and much more. In fact, over the years, research has established definite links between several medical and oral health conditions and in doing so have stressed the importance of a dental professional's role in early diagnosis or influencing the prognosis of medical conditions in a patient. Early diagnosis especially becomes important when considering that more than 120 medical conditions manifest first in the oral cavity (Rudman et al. 2010).

[1] Periodontal health refers to the health of the supporting structures of a tooth
[2] Inflammation of the periodontium (supporting structures of a tooth)

Therefore, the medical and oral health (M-OH) domains require decision support systems that can: i) integrate information from disparate sources, ii) unambiguously share information and communicate with other systems, and iii) perform automated decision tasks and provide explanations for the outputs. The following research is work in progress and it investigates how the semantic web tools, such as ontologies and rules, can be applied to connect the M-OH domains by developing a knowledge base that can be reused by the medical and dental information systems for semantic interoperability and reasoning for decision making. The contributions of this paper are threefold: a) developing a comprehensive cross-domain knowledge base that is generic enough to be reused by various health decision support applications, b) presenting an approach to achieve cross-domain communication between two theoretically inter-dependent but practically separate healthcare domains, and c) demonstrating the application of Semantic Web technologies to develop interoperable decision support systems.

The rest of the paper is organised as follows: related work is discussed next, followed by a brief discussion of semantic web technologies specifically those that we use in our work. In the approach section, we discuss in detail the design-science approach adopted by us and identify the challenges that arise during the various steps. With the help of example use cases, we demonstrate the working of our knowledge base. Finally, in the section on concluding remarks, we sum up the paper re-stressing upon the motivation for our work and subsequently its contributions.

## RELATED WORK

There have been ongoing efforts to build systems that can seamlessly share patient information between the medical and dental domains. Large-scale systems in the U.S. such as VistA, CattailsMD™, and the Indian Health Service Health Information System allow for such sharing of information (Rudman et al. 2010). However, unlike the decision support available over a patient's medical information in these systems, there is no such support provided over the shared information. This is primarily because these systems use reference terminologies that do not allow for complex reasoning tasks that decision support systems require. Thus, it is left to the practitioners to manually peruse the information and identify existing or possible associations, which may lead to information overload, oversight and defeat the purpose of sharing the information. Therefore, it is vital that the interdependencies between medical and oral conditions be consolidated and made available in a machine interpretable format for reuse and provision of decision support. Additionally, this information must be shareable between disparate systems so that patient information can be accessible as and when required by authorised health practitioners. Effectively, this sharing represents semantic interoperability, which is achievable by using semantic web tools and techniques such as ontologies. However, there is no comprehensive and representative ontology covering both medical and oral health concepts and consequently to the best of our knowledge, there are no automated decision support systems for reasoning over the shared medical and oral health information of patients. The most comprehensive and widely terminology system as of today is the Systematised Nomenclature of Medicine – Clinical Terminology (SNOMED-CT) (IHTSDO 2013) with 311,000 concepts, but it does not contain several terms and relationships that are important for both the medical and oral health domains (Goldberg et al. 2005). Moreover, the SNOMED-CT has several structural deficiencies that make reasoning over it for decision support tasks error-prone (Héja et al. 2008). Hence, the primary research challenges include developing: i) cross-domain systems capable of performing automated inferencing over the shared information, and ii) a reusable and shareable knowledge base containing formal, machine-interpretable, and standardised representations of interdependent medical and oral conditions.

## SEMANTIC WEB TECHNOLOGIES AND DECISION SUPPORT

### Ontology

Studer et al. (1998) defined ontology as "a formal, explicit specification of a shared conceptualization." A formal ontology thus ensures retention of meaning and accuracy of the information exchanged thereby enabling semantic interoperability between different systems. Besides, as opposed to terminologies and classification systems, which are static structures for knowledge reference, ontologies allow domain knowledge reference, reuse and reasoning (Noy and McGuinness 2001).

### Web Ontology Language

Web Ontology Language version 2 (OWL 2), which is a Web Standard, is an expressive ontology representation language for describing the semantics of knowledge (W3C 2013; w3schools 2012). It is based on the Description Logic (DL) *SROIQ*, which is decidable (Horrocks et al. 2006). However, OWL 2 is not decidable in its full form. Therefore, a subset OWL 2 DL, which is decidable is used for reasoning tasks and to take advantage of the various reasoners available (Hitzler et al. 2009). An OWL 2 ontology primarily consists of (Hitzler et al. 2009): i) axioms – the basic statements in an OWL ontology, ii) entities – the terms used for representing real world objects, and iii) expressions – these are complex descriptions derived from the combinations of various entities.

Further, any DL ontology has two main parts – TBox, which is the Terminological Box and ABox, which is the Assertional Box. The TBox contains the OWL class expression axioms such as subclass, equivalent class and disjointness, while the ABox contains OWL facts that is, the asserted individuals. For example, a TBox statement will be every oral infection is an infection and the corresponding ABox statement will be candidiasis is a type of oral infection.

## Semantic Web Rule Language

For the purpose of retaining decidability and classifying in polynomial time, there are several restrictions employed in OWL 2 thereby limiting its expressivity. For example, OWL 2 cannot express the relation *child of married parents* (Kuba 2012), which is basically a relation between individuals with which another individual is related. For such purposes, rules are used to enhance the expressivity of the underlying ontology language. Further, the rules provide actionable knowledge so that it is possible to develop decision support tasks in the form of alerts, reminders, recommendations, guidelines and diagnosis. However, in order to maintain semantic compatibility of the rules with the ontology, the rule language must be semantically compatible with OWL. The W3C proposal, Semantic Web Rule Language (SWRL) (Horrocks et al. 2004), provides a Horn clause rules extension to OWL in a semantically coherent manner. The basic structure of SWRL rule is of the form *antecedent* → *consequent* that is, if the antecedent or body of the rule is true then it is implied that the consequent or head is true as well and holds. The antecedent and consequent consists of a conjunction of atoms in the form $a_1 \wedge \ldots \wedge a_n$. Limiting the rule atoms to the named classes and properties within the base OWL ontology ensures interoperability of the ontological knowledge embedded within the rules with other OWL ontologies, which may or may not support SWRL (Horrocks et al. 2004). Moreover, such restrictions facilitate translation of SWRL rules to other rule systems such as Prolog, production rules and SQL and also improve tractability of the reasoning tasks that are performed over the rules (Horrocks et al. 2004). In this format thus, the previously mentioned relation *child of married parents* can be expressed as:

Person(?x) ∧ hasParent(?x, ?y) ∧ hasParent(?x, ?z) ∧ hasSpouse(?y, ?z) → ChildOfMarriedParents(?x)

where Person and ChildOfMarriedParents are named classes in the underlying OWL ontology; hasParent and hasSpouse are named properties; and ?x, ?y, ?z are variables. The rule states that a person whose parents are married is essentially a child of married parents.

## Reasoners

The OWL ontology and SWRL rules can thus together form a knowledge base[3] for a specific domain. In fact, OWL and rules are considered as the "foundations of semantic web technologies". An inference engine is then required to reason over this knowledge base to discover the hidden relationships. Some of the well-known reasoners include Pellet (Sirin et al. 2007), Hermit (Shearer et al. 2008), Fact++ (Tsarkov and Horrocks 2006), Kaon2 (Motik and Studer 2005), and RacerPro (Haarslev and Müller 2001) among others. We have selected Pellet for our work, which is based in Java and is available as open source. In addition to being a very efficient reasoner, the newer versions of Pellet provide native support for SWRL, albeit limited, thereby combining the knowledge for the reasoning process to provide more accurate and comprehensive results. Pellet also provides an explanation facility, which justifies the inferencing result by showing the pathways that were used to reach the specific decision.

With the Semantic Web technologies gaining increased maturity and with the increasing need for health systems to be interoperable, several biomedical ontologies such as SNOMED-CT, Foundational Model of Anatomy (FMA), Medical Subject Headings (MeSH) and National Cancer Institute (NCI) Thesaurus have been converted to OWL (Golbreich and Horrocks 2007). However, none of these ontologies use SWRL to extend the knowledge represented in the ontology and form a comprehensive knowledge base representing their respective domains.

# APPROACH

We apply the semantic web technologies namely OWL and SWRL to build our cross-domain knowledge base. Central to our approach are use cases that guide the entire development process, and are also used to develop competency questions to validate and evaluate the knowledge base for accuracy, consistency and comprehensiveness. Our approach is based on the design-science research principles as discussed in (Hevner et al. 2004). There are 4 main steps as discussed next. Figure 1 shows our developmental framework.

[3] For the purpose of this paper, knowledge base is referred to indicate ontology and rules together

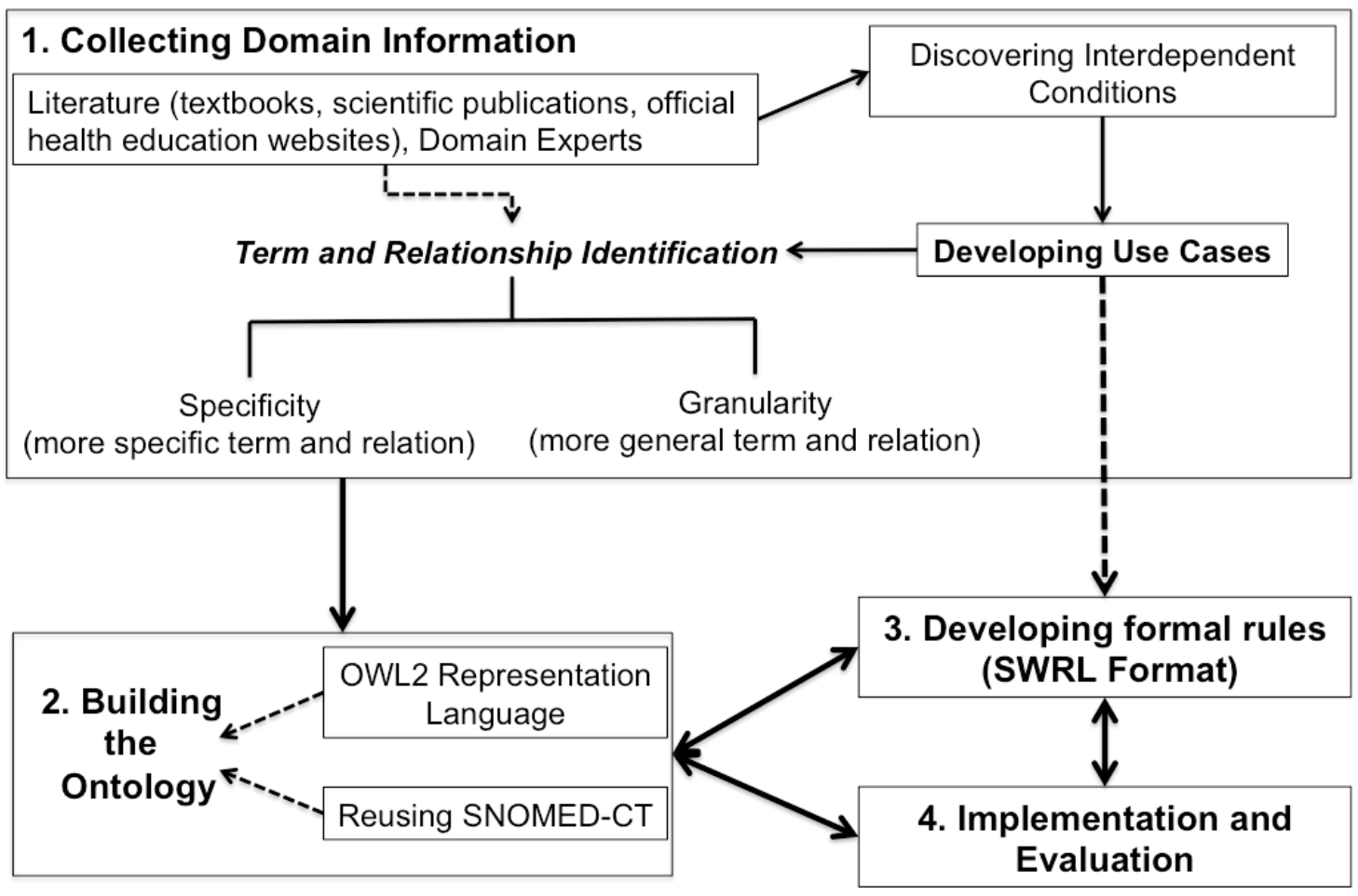

Figure 1: Developmental framework

## Step 1. Collecting Domain Information

The first task involves scoping that is, identifying the domain and its boundaries for representation. For this research, the domain has already been identified as the region that lays at the intersection of the medical and oral health domains and which represents the inter-dependent conditions from both the domains. These conditions are obtained from various sources including scientific literature and domain experts' knowledge. Three domain experts are involved in our development process of whom one is a general practitioner and the other two are dental surgeons and they are consulted to verify the correctness of the conditions and the rules formed from them. The conditions are developed into use cases, which then form our reference for discovering the terms to be modelled and the relationships between them, which are converted into object properties within the formal ontology in the later stages. Table 1 shows some example use cases obtained from literature, and the terms and relationships that are obtained from those. We will refer to use cases 3 and 5 throughout the rest of the paper.

Table 1. Example Use Cases and Terms and Relationships Derived From Them

| | Use Cases (made succinct here) | Terms | Relationships |
|---|---|---|---|
| **1.** | Candidiasis and Oral Hairy Leukoplakia are early indicators of the presence of Human Immunodeficiency Virus (Chattopadhyay et al. 2005; Laskaris et al. 1992). | Pseudomembranous Candidiasis, Oral Hairy Leukoplakia, HIV Infection | Early Indicator, Diagnostic |
| **2.** | The presence of periodontitis in pregnant women has been associated with the birth of low birth weight infants (Li et al. 2000). Therefore, maintenance of good oral hygiene or providing periodontal treatment is essential during pregnancy. | Pregnant, Low Birth Weight Infant, Periodontitis, Periodontal Therapy, Good Oral Hygiene | Patient At Risk, To Maintain, Recommended Therapy, Preventive Measure |
| **3.** | The progress of Diabetes Mellitus (DM) is adversely affected by periodontal disease. Conversely, poorly controlled DM exacerbates periodontal disease (Mealey 2006). Therefore, a patient with either of these conditions must be managed collaboratively by the medical and oral health practitioners. | Diabetes Mellitus, Diabetes Mellitus Type 2, Periodontal Disease, Periodontitis | Affects, Influences, Interacts With |
| **4.** | An untreated periodontal abscess can lead to the development of Ludwig's Angina, which if left untreated, can cause fatal complications such as asphyxia (Marcus et al. 2008). | Periodontal Abscess, Untreated/No Treatment, Ludwig's Angina, Asphyxia | Causes, Leads To, Has Complication |
| **5.** | If a patient with any form of congenital heart disease (CHD) and poor oral hygiene undergoes surgical dental extraction, then the resulting transient bacteraemia will most likely react with the underlying CHD and put the patient at risk of a bacterial endocarditis. Antibiotic prophylaxis is recommended to prevent the occurrence of endocarditis in such cases (Li et al. 2000; Roda et al. 2008). | Congenital Heart Disease, Poor Oral Hygiene, Surgical Extraction of Tooth, Bacteraemia, Bacterial Endocarditis, Antibiotic, Prophylaxis, Antibiotic Prophylaxis | Has Condition, Undergoes Procedure, Causes, Patient At Risk, Preventive Recommendation |

These terms are analysed for lexical and semantic similarities and differences. Each term is further refined to derive a more general and a more specific term. This way, more terms are discovered and an initial hierarchy is obtained. Neighbourhood terms are further identified from the associations obtained from literature. The main research challenge in this step is to discover all possible terms to represent comprehensiveness of both the domains and at the same time, all the terms must be relevant to both the domains. This ensures that only necessary terms are modelled in the ontology and overloading of information does not happen.

## Step 2. Building the Cross-Domain Ontology

Since the ontology represents cross-domain M-OH knowledge, we have named it Oral-Systemic[4] Cross-domain Ontology (OSHCO). In this step, the identified terms and relationships from the previous steps are matched with the corresponding concepts and properties in our reference terminology, SNOMED-CT and the appropriate hierarchical structure within the reference terminology identified. However, not all the required terms and relations are available in SNOMED-CT. In that case, a more general or specific term is identified and the hierarchy that is the closest match to the context of the required term is selected and the term is added to it. The reader is referred to (Shah et al.) for a detailed description of reusing SNOMED-CT. As discussed previously, the ontology is represented in OWL 2. This is to ensure semantic interoperability between the systems using our ontology with other systems that use the corresponding Web Standards. A significant research challenge at this step is to ensure that in modelling OSHCO as closely as possible to SNOMED-CT, the structural and modelling pitfalls of the latter are not replicated in the resulting ontology. This is important because OSHCO being a cross-domain ontology contains a rich density of relationships to represent the various use cases correctly and in doing so it is extremely easy to convert into a heavy ontology with a large number of terms and properties thereby making it practically inconvenient to run and reason over on local machines for real-time decision support tasks, which is one of the major issues with SNOMED-CT (Dentler et al. 2011).

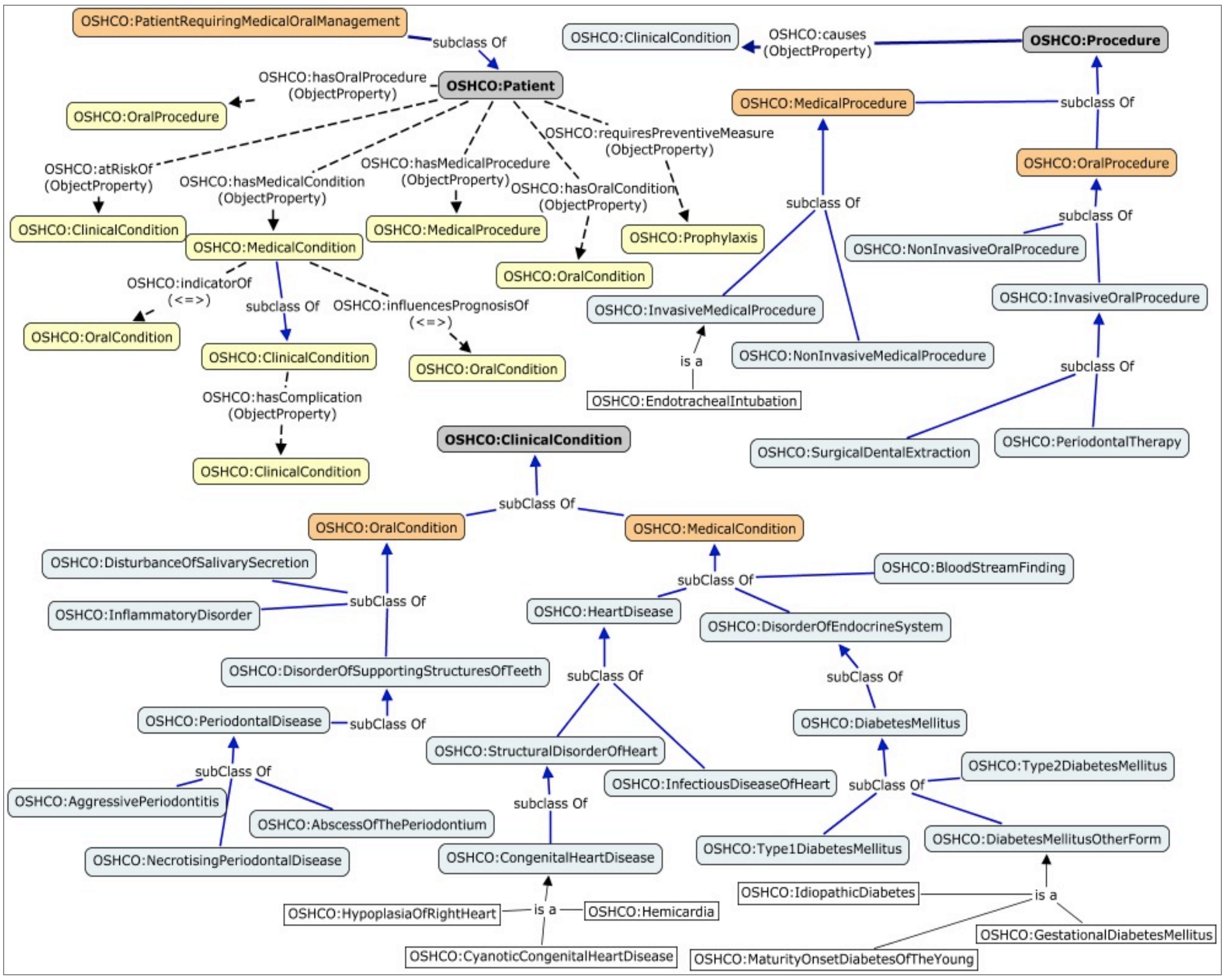

Figure 2: A portion of OSHCO

We used Protégé 4.2 (Stanford 2013), an open source ontology editor to build and validate our ontology. Figure 2 shows a portion of our ontology including classes, subclasses, named individuals and object properties as built in Protégé and exported into CMap[5]. Three main classes namely patient, procedure and clinical condition with some of their subclasses, as well as few relationships (referred to as object properties in OWL) for the patient class can be seen. The properties have been modelled according to the use cases to connect the inter-dependent conditions thereby linking the M-OH domains within the ontology. Moreover, as we will discuss in the next section, the relationships of patient class to the other classes help in deriving actionable knowledge from the asserted facts.

[4] Medical conditions are also referred to as Systemic conditions

## Step 3. Developing Formal Rules

The use cases from step 1, and the terms and properties (relationships) that are used in the ontology in step 2 serve as the blueprint for writing formal rules in SWRL at this stage. By using only the named ontology classes, we ensure that our rules can be translated to different rule formats, are interoperable with other OWL based ontologies that may or may not support SWRL, and that the rules remain decidable. We develop rules in two situations – where actionable knowledge is required and where conditions cannot be expressed in OWL. For example, with respect to use case 3 described in table 1, it is possible to express in our OWL based ontology OSHCO that if a patient has some form of DM and periodontal disease then the patient should be automatically classified into a new class of patients who require collaborative (medical-oral) management:

$$\text{PatientRequiringMedicalOralManagement} \equiv \text{Patient} \sqcap (\exists\text{hasMedicalCondition.DiabetesMellitus} \sqcap \exists\text{hasOralCondition.PeriodontalDisease})$$

However, it is not possible to express that a patient should be classified into the class of patients who require collaborative management only if he/she has those medical and oral conditions that are interdependent or in other words, influence each other's prognosis. However, it can be expressed in SWRL as shown in Table 2, rule complex 1. This is an example that shows how rules can add to the expressivity of the ontology. Rule complex 2 on the other hand shows how rules add actionable knowledge to the ontology. The rules in complex 2 warn the user of what is likely to happen if certain conditions are met and what preventive measure is recommended in such cases.

Table 2. Rule Complexes Represented in SWRL Format

| | |
|---|---|
| Rule Complex 1 | Patient(?x) ∧ hasMedicalCondition(?x, ?y) ∧ hasOralCondition(?x, ?z) ∧ MedicalCondition(?y) ∧ OralCondition(?z) ∧ hasInterdependency(?y, ?z) → PatientRequiringMedicalOralManagement(?x) |
| | MedicalCondition(?y) ∧ OralCondition(?z), influencesPrognosisOf(?y,?z) → hasInterdependency(?y, ?z) |
| Rule Complex 2 | Patient(?x) ∧ hasOralCondition(?x, PoorOralHygiene) ∧ hasOralProcedure(?x, ?y) ∧ SurgicalDentalExtraction(?y) → atRiskOf(?x, BacteraemiaDueToSurgicalDentalProcedure) |
| | Patient(?x) ∧ atRiskOf(?x, ?z) ∧ hasMedicalCondition(?x, ?y) ∧ Bacteraemia(?z) ∧ CongenitalHeartDisease(?y) → atRiskOf(?x, BacterialEndocarditis) |
| | Patient(?x) ∧ atRiskOf(?x, BacterialEndocarditis) → requiresPreventiveMeasure(?x, AntibioticProphylaxis) |

Figure 3 is a snapshot of Protégé showing rule complexes 1 (for example patient 'Tim') and 2 (for example patient 'Sam') and the corresponding inferences derived from these rules. The expressions coloured in yellow and within the dotted lines are the new inferences obtained after reasoning while the rest are asserted statements and facts. As mentioned before, Pellet also provides a justification for the output by showing the path that led to that specific output. Figure 4 also shows one of four justification paths traversed by Pellet for rule complex 1. The justification module is especially important since any errors in the output can be traced to their source by referring to the path traversed by the reasoner and changes can be made during the development process itself.

## Step 4. Implementation and Evaluation

In line with the design-science approach, our proposed approach is iterative. Accordingly, we perform regular validation using Pellet to check for ontology consistency, concept satisfiability, classification, and realisation using Pellet. Moreover, the domain experts are also consulted regularly to check the correctness of the represented concepts and rules. In addition to regular ontology validation, we performed preliminary evaluation by querying over the knowledge base, as discussed next.

[5] http://ftp.ihmc.us/

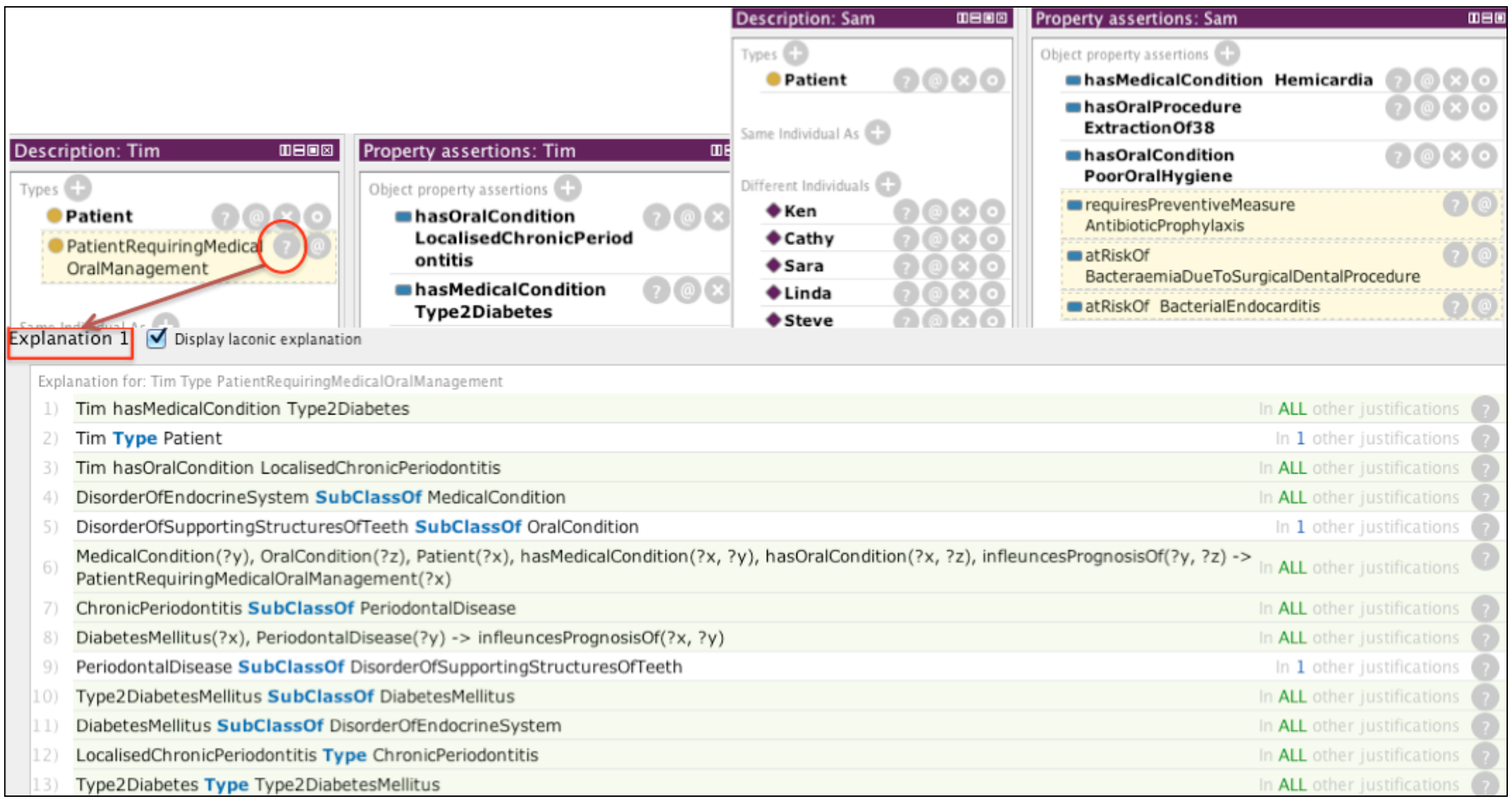

Figure 3: Snapshot of example inferencing outputs and reasoning justification in Protégé

Use of competency questions for ontology validation and evaluation is well-established in literature (Noy and McGuinness 2001). Accordingly, for every use case, we developed simple and complex competency questions with the guidance of domain experts such that the results: i) validate OSHCO – classification correctness, consistency, completeness and specificity, ii) validate rules – correctness and completeness, and iii) compare the performance of OSHCO with and without rules. Following are example questions for use cases 3 and 5 mentioned in Table 1.

Use Case 3:

- Question 1 (complex) – Identify all the patients who have periodontal disease and identify medical conditions whose prognosis is affected by periodontal disease.
- Question 2 (complex) – Identify patients who have both, a type of Diabetes Mellitus and periodontal disease and list the types they have.
- Question 3 (simple) – Identify patients require collaborative medical-oral monitoring.
- Question 4 (simple) – If patient 'xyz' requires collaborative medical-oral monitoring, then which underlying medical and oral conditions does the patient have?

Use Case 5:

- Question 1 (complex) – Identify all the patients who are at a risk of developing bacteraemia due to dental extraction and identify the underlying medical conditions they have.
- Question 2 (complex) – What conditions is patient 'xyz' at risk of developing and what preventive measures, if any, are required?
- Question 3 (simple) – Which patients require antibiotic prophylaxis as a preventive measure?

For the purpose of querying, we manually added simulated patient cases to our knowledge base. Thereafter, using the query language, SPARQL (Harris and Seaborne 2010), we questioned our knowledge base for answers. Shown below is an example query for use case 3 question 2 and a snapshot of the output on running the query in protégé is shown in figure 4 (details are omitted for brevity).

```
PREFIX rdf: <http://www.w3org/1990/02/22-rdf-syntax-ns#>
PREFIX oshco: <http://www.semanticweb.org/ontologies/2013/6/OSHCO.owl#>
SELECT ?Patient ?TypeOfDiabetesMellitus ?TypeOfPeriodontalDisease
  WHERE {?Patient  oshco:hasMedicalCondition  ?TypeOfDiabetesMellitus ;
          oshco:hasOralCondition  ?TypeOfPeriodontalDisease ;
  rdf:type oshco:Patient .
 ?TypeOfDiabetesMellitus rdf:type oshco:DiabetesMellitus .
 ?TypeOfPeriodontalDisease rdf:type oshco:PeriodontalDisease . }
```

Results

| ?Patient | ?TypeOfDiabetesMellitus | ?TypeOfPeriodontalDisease |
|---|---|---|
| oshco:Steve | oshco:DrugInducedDiabetes | oshco:AcuteNecrotisingUlcerativePeriodontitis |
| oshco:Tim | oshco:Type2Diabetes | oshco:LocalisedChronicPeriodontitis |
| oshco:Ken | oshco:Type2Diabetes | oshco:PeriodontalAbscess |
| oshco:Sara | oshco:MaturityOnsetDiabetesOfTheYoung | oshco:MarginalPeriodontitis |
| oshco:Martin | oshco:PreDiabetes | oshco:GeneralisedAggressivePeriodontitis |
| oshco:Cathy | oshco:GestationalDiabetesMellitus | oshco:GeneralisedAggressivePeriodontitis |
| oshco:Linda | oshco:ImmuneMediatedDiabetes | oshco:CombinedPeriodonticEndodonticLesion |

Figure 4: Snapshot of the results for use case 3 query 2

## Initial Results

The evaluation metrics and results for the 5 use cases mentioned in step 1 are given in table 3. For each use case, a minimum of 5 questions were queried. We divided each question into simple and complex and executed the corresponding queries twice – first with OSHCO alone, that is based only on the expressions modelled in the base ontology without addition of SWRL rules; and thereafter with rules. Thus, for each type of question, three results are obtained – questions resolved by OSHCO alone, resolved by OSHCO and rules together (OSHCO+R) and those left unresolved (UR). The results are summarised in figure 5.

Table 3. Initial Results

| | **Simple Questions** | | | **Complex Questions** | | |
|---|---|---|---|---|---|---|
| ***Use Case*** | **OSHCO** | **OSHCO+R** | **UR** | **OSHCO** | **OSHCO+R** | **UR** |
| *1* | 4 | 5 | 1 | 0 | 4 | 0 |
| *2* | 2 | 5 | 0 | 3 | 6 | 1 |
| *3* | 1 | 3 | 1 | 2 | 4 | 1 |
| *4* | 4 | 8 | 0 | 1 | 2 | 2 |
| *5* | 1 | 2 | 1 | 0 | 5 | 1 |

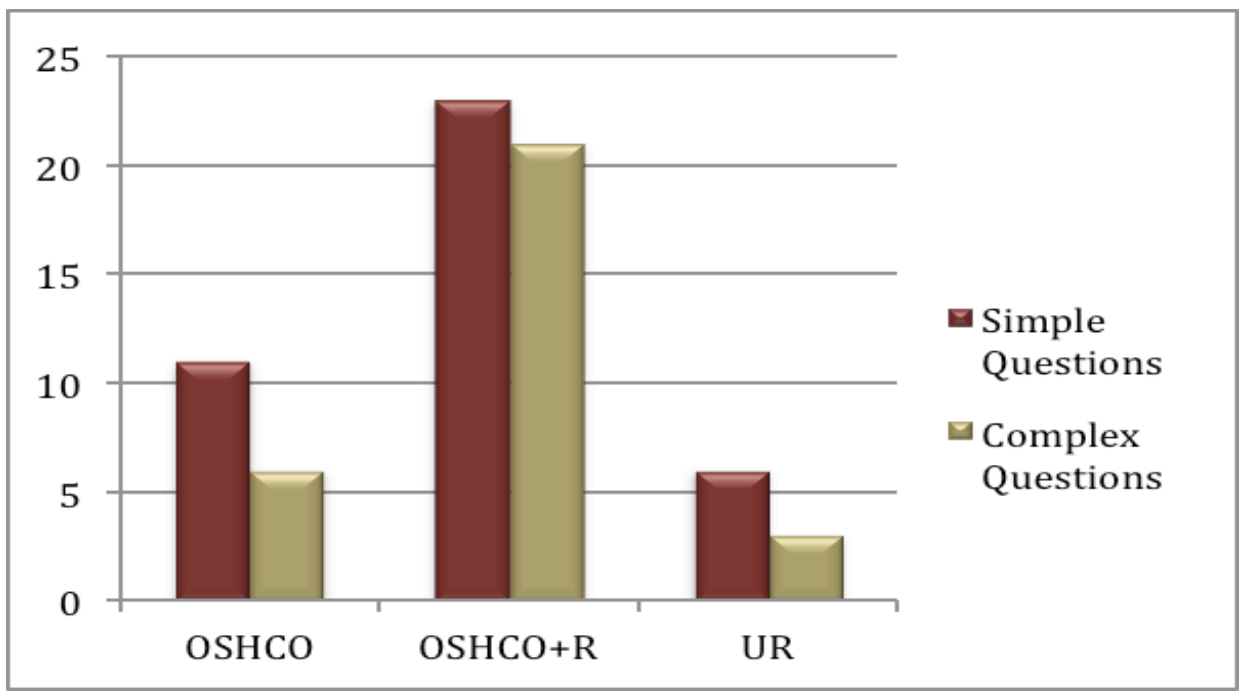

Figure 5: Summary of Initial Results

## Discussion

By executing each query twice, with OSHCO alone and with rules, we could validate both the ontology and rules and also identify those questions that could not have been answered without rules. Essentially, it helped us to validate our approach of developing a cross-domain ontology specific to the M-OH domains and using rules with the ontology. From the results it can be seen that a higher number of questions, 62, were resolved as against only 8 that remained unresolved. This shows that developing a comprehensive ontology for the M-OH domains is essential to address a large number of use cases that contain oral health concepts, which would not have been possible with any of the existing terminologies since none of them contain all the oral health concepts. Moreover, as predicted, OSHCO with rules was able to answer 21 questions for each complexity type, which is significantly higher than the 6 questions answered by OSHCO alone. However, OSHCO alone resolved more number of simple questions than complex; this is again as expected since representing complex situations in OWL is not possible. The two outcomes above validate our argument as presented in step 3 that OWL has several limitations in expressivity and generating actionable knowledge; and this can be addressed to a great extent by adding rules. As our next step, we will implement our knowledge base in a prototype system for rigorous evaluation to ensure that the system would perform similarly with actual patient cases and in real-world settings.

## Limitations

As can be seen from figure 5, there are a few questions that remained unresolved and the reasons can be attributed to some well-known limitations of OWL, one of which is monotonicity. In plain terms, it means that addition of new information does not change or alter pre-existing information. Consequently, OWL does not support negation as failure and closed world assumption. In other words, OWL is based on open world assumption that is, if something is unknown then OWL treats it as unknown unlike closed world models that would treat it as false. Therefore, negation as failure is not supported either that is, a question such as "identify patients who have DM but do not have periodontal disease" is not possible unless it is explicitly stated so for the patients who do not have periodontal disease, otherwise OWL assumes that the patient(s) may have the condition but we are not aware of it. Adding SWRL rules does not address these issues either, because they themselves are an extension of the OWL logic and hence make the same assumptions that OWL does. These limitations are being increasingly investigated by semantic web researchers and some solutions such as description logic programs have been proposed to address them (Motik et al. 2007). In our future work, we will be looking at these solutions and analyse how they can be employed to improve the decision making capabilities of our cross-domain knowledge base.

# CONCLUDING REMARKS

There are very few health information systems that enable sharing of medical and oral health information of patients and those are based on standards, terminologies and/or classification systems. Therefore, retention of meaning and reasoning over the shared information has not been possible that is, this sharing is limited to data integration and has not progressed to semantic interoperability. Moreover, the heterogeneity in data collection and storage formats across the two domains has further restricted meaningful information sharing that can be converted into practical benefits. In this research we have attempted to address the above problems using Semantic Web technologies such as formal ontologies and rules. For the same, we have discussed our development approach in detail and provided a brief description and justification for each technology that we use in our approach. Our approach is novel in that, to the best of our knowledge, there are not many examples in healthcare and especially none that make use of OWL ontology and SWRL rules to develop a reusable and comprehensive knowledge base to bring together two domains that are otherwise functioning in isolation for most practical purposes. We therefore envision that our approach will be generic enough to guide the development of various cross-domain healthcare applications such as decision support systems.

This paper has significant novel contributions in that it harnesses the semantic web technology in direct and important ways to enable the construction of a smarter and humane service for patients and healthcare practitioners – an area that will affect and improve the lives of everyone involved. The solutions developed in this research can make it possible for medical and oral health care practitioners to query and conduct automated inference over massive and distributed data, discover new knowledge and generate novel hypotheses in health management systems. The research also has the potential to become a very important demonstration of the power of integrating World Wide Web, pervasive computing, data processing and data mining technologies with health care.